\documentclass[11pt,a4paper]{article}
\usepackage[hyperref]{acl2020}
\usepackage{times}
\usepackage{latexsym}
\usepackage{tcolorbox}
\usepackage{bbding}
\usepackage{url}
\usepackage{multirow}
\usepackage{amssymb}
\usepackage{amsmath}
\usepackage{amsfonts}
\usepackage{multirow}
\usepackage{booktabs}
\usepackage{graphicx}
\usepackage{url}
\usepackage[toc,page]{appendix}

\usepackage{tcolorbox}
\usepackage{setspace}
\newcommand{\dataname}{\textsc{M}u\textsc{T}ual}
\definecolor{babyblue}{rgb}{0.54, 0.81, 0.94}
\definecolor{ballblue}{rgb}{0.13, 0.67, 0.8}
\definecolor{bluegray}{rgb}{0.4, 0.6, 0.8}
\usepackage{microtype}

\aclfinalcopy 
\def\aclpaperid{1420} 

\title{\dataname: A Dataset for Multi-Turn Dialogue Reasoning}

\author{
  Leyang Cui$^{\dag \ddag}$\thanks{\ \  Contribution during internship at MSRA.}, Yu Wu$^\Diamond$, Shujie Liu$^\Diamond$, Yue Zhang$^{\ddag}$, Ming Zhou$^\Diamond$ \\
  $^\dag$Zhejiang University \\
  $^\Diamond$Microsoft Research Asia \\
  $^\ddag$School of Engineering, Westlake University \\
  $^\ddag$\{cuileyang,zhangyue\}@westlake.edu.cn\  
  $^{\Diamond}$\{Wu.Yu,shujliu,mingzhou\}@microsoft.com \\
  }

\date{}

\begin{document}
\maketitle

\begin{abstract}

Non-task oriented dialogue systems have achieved great success in recent years due to largely accessible conversation data and the development of deep learning techniques. Given a context, current systems are able to yield a relevant and fluent response, but sometimes make logical mistakes because of weak reasoning capabilities. To facilitate the conversation reasoning research, we introduce MuTual, a novel dataset for \textbf{Mu}lti-\textbf{Tu}rn di\textbf{al}ogue Reasoning, consisting of 8,860 manually annotated dialogues based on Chinese student English listening comprehension exams. Compared to previous benchmarks for non-task oriented dialogue systems, MuTual is much more challenging since it requires a model that can handle various reasoning problems. 
Empirical results show that state-of-the-art methods only reach 71\%, which is far behind the human performance of 94\%, indicating that there is ample room for improving reasoning ability. 
\dataname\ is available at  \url{https://github.com/Nealcly/MuTual}.
\end{abstract}
\section{Introduction}
Building an intelligent conversational agent is one of the longest running goals in AI. 
Existing conversational agents can be categorized into task-oriented dialogue systems \cite{smartreply} and non-task-oriented chatbot systems \cite{xiaoice, Wu_2019}. 
Owing to the rise of deep learning techniques and the large amount of conversation data for training \cite{ubuntu-dataset, douban-dataset, e-commerce-dataset}, we are now witnessing promising results of chatbots both in academia and industry \cite{pan-etal-2019-improving, IoI}.

\begin{figure}
    \centering
    \includegraphics[width=0.5\textwidth]{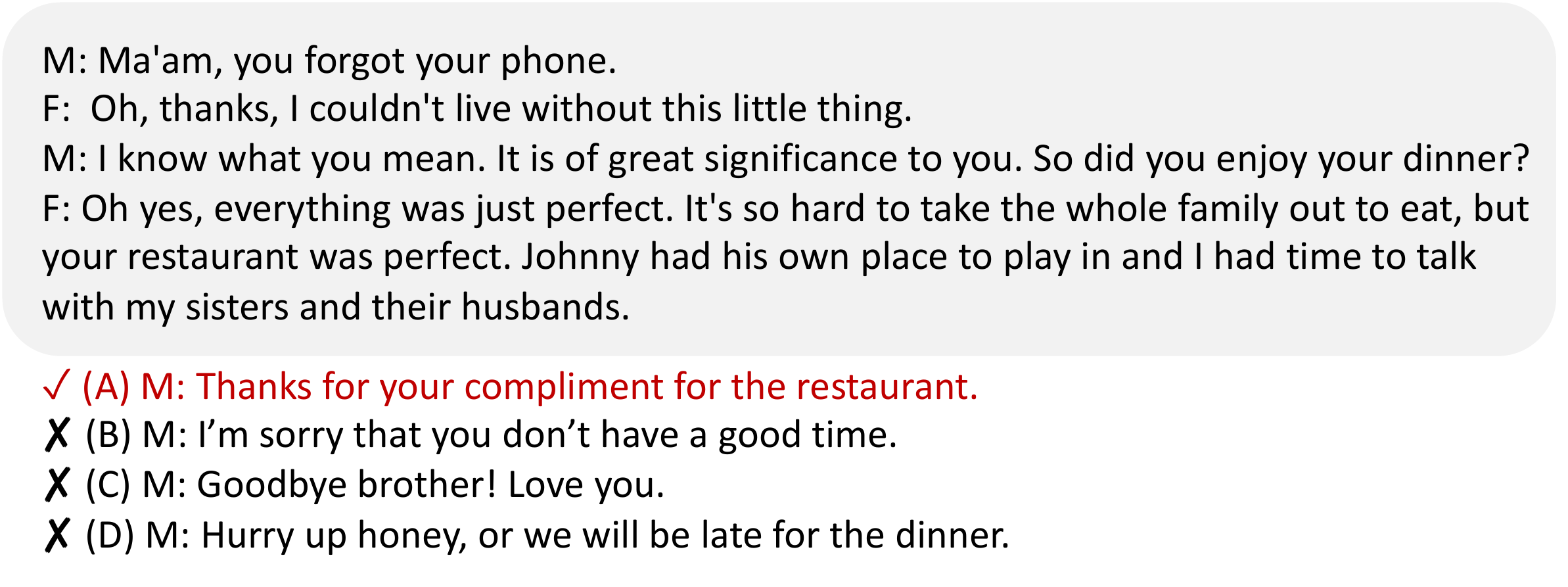}
    \caption{B is incorrect because there is no reason to apologize. C and D can be excluded because the relationship between two speakers are waiter and customer based on the context.}
    \label{fig:example-dataset}
    \vspace{-10pt}
\end{figure}

Neural dialogue systems are trained over a large dialogue corpus and used to predict responses given a context. There are two lines of methods. Retrieve-based methods and generation based methods rely on matching scores and perplexity scores, respectively. 
Due to the development of text matching and pre-training models \cite{bert,roberta}, a machine is able to achieve highly competitive results on these datasets, even close to human performance. For instance,  ESIM \cite{esim} achieves 88$\%$ on the Dialogue NLI \cite{dialogue-nli}, and BERT achieves 85.8\%, 93.1\% and 98.5\% in terms of R$_{10}@1$,  R$_{10}@2$ and  R$_{10}@5$ on the Ubuntu Corpus \cite{bert-result}. 

\begin{table*}[t] \small
    \centering
    \begin{tabular}{c|cccc}
    \hline
    dataset & Task & Reasoning & Domain & Manually\\
    \hline
    Ubuntu \cite{ubuntu-dataset} & {\bf Next Utterances Prediction} & \XSolidBold & Technique & \XSolidBold \\
    PERSONA-CHAT \cite{personalizing-dataset} & {\bf Next Utterances Prediction} & \XSolidBold & Persona & \CheckmarkBold \\
    Dialogue NLI \cite{dialogue-nli} & {\bf Next Utterances Prediction} & \XSolidBold & Persona & \XSolidBold \\
    CoQA \cite{coqa} & Conversational QA & \CheckmarkBold & Diverse & \CheckmarkBold  \\
    Douban \cite{douban-dataset} & {\bf Next Utterances Prediction} & \XSolidBold & Open & \XSolidBold \\
    DREAM \cite{dream} & Reading Comprehension & \CheckmarkBold & Open & \CheckmarkBold \\
    \hline
    WSC \cite{wsc-dataset} & Coreference Resolution & \CheckmarkBold & Open & \XSolidBold \\
    SWAG \cite{swag-dataset} & Plausible Inference & \CheckmarkBold & Movie & \XSolidBold\\
    CommonsenseQA \cite{commonsenseqa} & Reading Comprehension & \CheckmarkBold & Open & \CheckmarkBold \\
    RACE \cite{lai-race} & Reading Comprehension & \CheckmarkBold & Open & \XSolidBold \\
    ARC \cite{ai2-dataset} & Reading Comprehension & \CheckmarkBold & Science & \XSolidBold \\
    DROP \cite{drop} & Reading Comprehension & \CheckmarkBold & Open & \XSolidBold \\
    Cosmos \cite{Cosmos} & Reading Comprehension & \CheckmarkBold & Narrative & \CheckmarkBold \\
    \hline
    \dataname & {\bf Next Utterances Prediction} & \CheckmarkBold & Open & \CheckmarkBold \\
    \hline
    \end{tabular}
    \caption{Comparison between our dataset and other datasets. { ``Manually'' indicates that human writing of the question or answers is involved in the data annotation process, rather than mere manual selection of data.}}
    \label{tab:compare}
    \vspace{-10pt}
\end{table*}

However, there is still a huge gap between high performance on the leader-board and poor practical user experience.
Chatbot engines often generate responses that are logically incorrect or violate commonsense knowledge \cite{xiaoice}. 
A likely reason is that current dialogue systems do not have strong reasoning skills, and most of the cases in previous benchmarks can be tackled by linguistic information matching. Previous work has demonstrated that neural encoders capture a rich hierarchy of syntactic and semantic information \cite{bert-analysis, bert-attention}. However, reasoning capability and commonsense knowledge are not captured sufficiently \cite{dialogue-commonsense}.

One important research question is how we can evaluate reasoning ability in chatbots, which can potentially allow us to bridge the gap between high performance on leader-board and unsatisfactory practical performance. To this end, we develop an open domain {\bf Mu}lti-{\bf Tu}rn di{\bf al}ogue reasoning dataset (\dataname) to facilitate conversation model reasoning capabilities. In particular, given a context, we prepare four response candidates, each of which is relevant to the context, but only one of them is logically correct. As shown in Figure~\ref{fig:example-dataset}, all responses follow the same topic, but only the first one is appropriated. 
It requires reasoning ability on social etiquette and relationship to make the correct choice, which is not considered by existing dialogue benchmarks.
 
We build our dataset based on Chinese high school English listening comprehension test data, where students are excepted to select the best answer from three candidate options, given a multi-turn dialogue and a question.
The original data is formatted as $\langle$dialogue, question, answer$\rangle$, which is not directly suitable for our goal since chatbots only concern about how to respond contexts instead of answering an additional question. Therefore, we ask human annotators to rewrite the question and answer candidates as response candidates. Then our dataset follows the traditional response selection setting \cite{ubuntu-dataset},  where a model should recognize a correct response from others for a multi-turn dialogue. 

The resulting dataset, \dataname, consists of 8,860 challenge questions, in terms of almost all questions involving reasoning, which are designed by linguist experts and high-quality annotators. We evaluate state-of-the-art retrieval-based models and pre-training models on \dataname.
The best method gives a R$@1$ of 71\%, which significantly underperforms human performance (94\%).
To the best of our knowledge, \dataname\ is the first human-labeled reasoning-based dataset for multi-turn dialogue.
We provide detailed analysis to provide
insights into developing potentially reasoning-based chit-chat dialogue systems.
\section{Related work}
\label{sec:relatedwork}

Table~\ref{tab:compare} compares our dataset with prior dialogue and reasoning related benchmarks. 

\textbf{Dialogue:}
The Ubuntu Dialogue Corpus is a large retrieval-based dataset \cite{ubuntu-dataset}, extracted from Ubuntu chat logs.
PERSONA-CHAT \cite{personalizing-dataset} considers consistent personality in dialogue. Crowd workers are required to act the part of a given provided persona, and chat naturally. 
Dialogue NLI \cite{dialogue-nli} is a natural language inference dataset modified from PERSONA-CHAT. It demonstrates that NLI can be used to improve the consistency of dialogue models.
CoQA \cite{coqa} is collected by pairing two annotators to chat about a passage in the form of questions and answers. 
Each question is dependent on the conversation history. There are also several large-scale datasets in Chinese, such as Sina Weibo \cite{weibo-dataset},  Douban Conversation Corpus \cite{douban-dataset} and  E-commerce Dialogue Corpus \cite{e-commerce-dataset}. 

\begin{figure*}[!t]
\centering
\includegraphics[width=\textwidth]{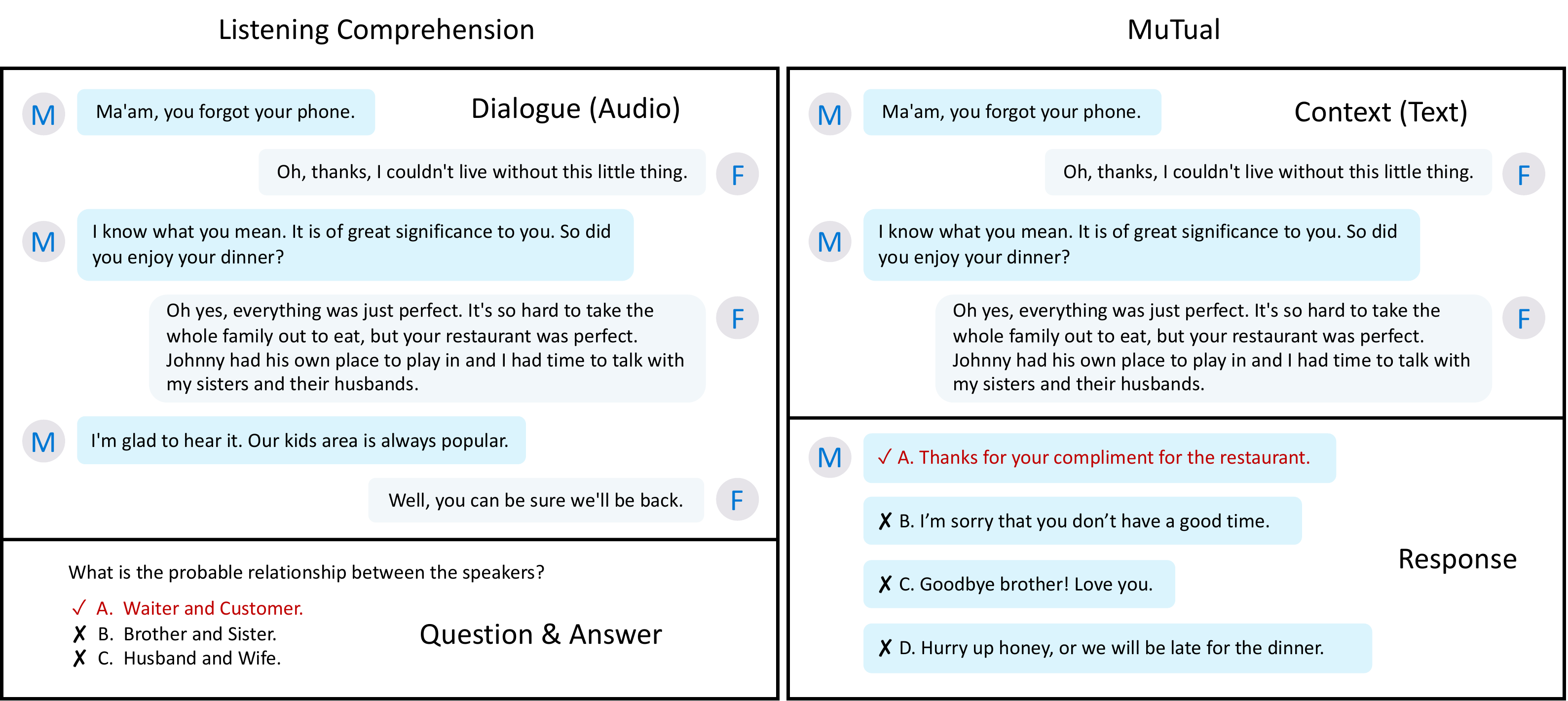}
\caption{The process of modifying the listening comprehension test data. 
}
\label{fig:data-construct}
\end{figure*}


As shown in Table~\ref{tab:compare}, most of the existing conversation benchmarks do not focus on testing reasoning ability. One exception is CoQA, which considers pragmatic reasoning. The difference is that CoQA is a machine comprehension dataset, in which conversations are based on a given passage. 
Another related reading comprehension dataset is DREAM \cite{dream}, which is designed specifically for challenging dialogue-based reading comprehension. It relies on an external question to test the model's understanding capability.
In contrast to the above dataset, our dataset is a next utterance prediction task, which is the fundamental problem in retrieval-based chatbots. In addition, our dataset requires various specific reasoning abilities, such as algebraic reasoning, intention prediction and so on, which is the main characteristic of our dataset.  



\textbf{Reasoning:}
Recently, efforts have been made to develop benchmarks and tasks to address reasoning for language understanding.
Winograd Schema Challenge \cite{wsc-dataset} is a reasoning-based coreference resolution task. Each pair of sentences differs by only one phrase. 
SWAG \cite{swag-dataset} is derived from pairs of consecutive video captions, including 113k short context each with four candidates endings. CommonsenseQA \cite{commonsenseqa} is a question answering dataset extracted from CONCEPTNET \cite{conceptnet}. Utilizing CONCEPTNET to construct the dataset ensures that questions directly target commonsense reasoning.
RACE is a machine reading comprehension dataset collected from English exams for Chinese students. 
AI2 Reasoning Challenge \cite{ai2-dataset} contains 7,787 genuine grade-school level science questions with a corpus of 14M science reference sentences. DROP \cite{drop} and COSMOS \cite{Cosmos} focus on factual understanding and commonsense comprehension, respectively. 

Despite their success, these datasets can hardly help chatbots directly. Following the traditional dialogue response selection setting, we deeply modify English listening comprehension conversation to form an utterance prediction task.




\section{Dataset}
\subsection{Collection}
\label{sec:collection}
The original listening comprehension materials and question-answer pairs are designed by linguist experts. Students are required to choose the best answer from three options for a question based on a piece of audio. To ensure students fully understand the audio, most of the questions need to be answered with reasoning capability.

We crawled the listening exams from public websites\footnote{All the problems in our dataset are freely accessible online without copyright by consulting the legal adviser.}. Since the audio is either a conversation between two people or a simple passage, we only crawled data in the conversation format. The raw data is formatted as triples $\langle$Conversation (audio), Question and Choices (text), Answer (image)$\rangle$. The following data pre-processing methods are applied to convert raw data to data in Figure \ref{fig:data-construct}.

\textbf{Step 1 Pre-processing:} If question and candidate choices in two problems are the same, we consider them as duplicates and delete one of them. If there are more than three candidate options in one problem, we randomly drop incorrect options until three candidates are left.

The answers are stored as images. We apply a commercial OCR system to convert images to text. It is easy to recognize the printed alphabet answer for the OCR system. We manually correct all OCR outputs to ensure quality.
In the original listening comprehension test, the conversation is stored as audio. We adopt a commercial ASR system to convert speech to text, and further recruit experienced annotators to correct the transcription errors. To further ensure the quality of the transcripts, they are double-checked by annotators in the next step. 
 
\textbf{Step 2 Candidate Response Creation:}  Figure~\ref{fig:data-construct} illustrates the process of modifying the listening comprehension problem. 
At first, an annotator is required to segment the original conversation, after clues to answer the question have appeared. Then, they construct positive response (Response A in Figure~\ref{fig:data-construct}) and negative responses (Response C and Response D) by consulting correct choice (Choice A) and incorrect choices (Choice B and Choice C), respectively. 
 To make MuTual more challenging, we further ask the annotator to construct one more negative response (Response B) based on the correct choice.
Through these steps, MuTual not only keeps the reasoning test designed by experts, but also introduces one more another type of reasoning for each instance. As shown in Figure~\ref{fig:data-construct}, Response C and D can be excluded based on the relationship between two speakers. But B is incorrect due to the attitude reasoning.

It is worth noting that all negative responses are logically correct if the context is not considered, but they are not appropriated responses if the context is taken into account. Therefore, our dataset focuses on multi-turn conversation reasoning rather than the logic of a sentence.
When framing a {\it negative response}, we encourage annotators to copy some phrases in the context to discourage a model that can solve the problem by text matching. We further calculate the {\it lexical overlap} between response and context. There are 9.98\% (10.63\%) words in the positive (negative) response that occur in the corresponding context, suggesting that MuTual is hard to solve by plain text matching.

Annotators in Step 2 are all English-major graduate students in Chinese, who are familiar with English language exams in China and fluent in English (pass the TEM-8\footnote{The highest level test for English majors as a foreign language in China.}). Annotators are required to draft annotate 170 instances repeatedly, until their labeling is sufficiently accurate to provide useful annotation. Because not all conversations are adapted to construct a reasoning-based response problem, the annotator has the right to skip the conversation.
We employ five annotators to construct the response, and two quality inspectors to check it. We discard the instance when inspectors doubt the uniqueness or correctness of the answer.

\begin{figure*}[!t]
    \centering
    \includegraphics[width=\textwidth]{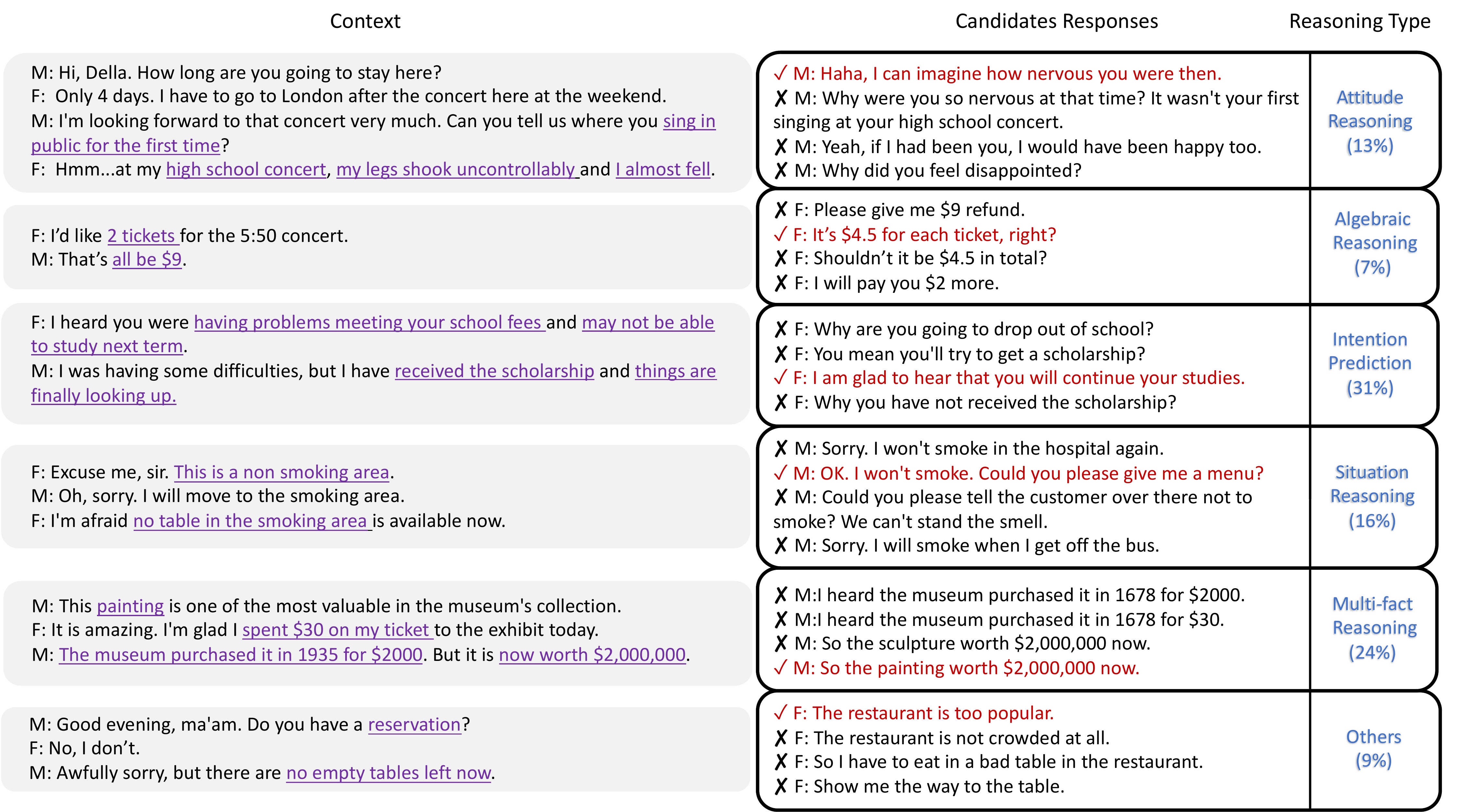}
    \caption{Examples from the MuTual dataset. All choices are relevant to context, but only one of them is logic correct. Some negative choices might be reasonable in extreme cases, but the positive one is the most appropriate.  Clue words are purple and underline. More examples are shown in Appendix A.}
    \label{figure:case}
\end{figure*}

\subsection{Analysis}
\label{sec:data analysis}

\begin{table}[t] \small
    \centering
    \begin{tabular}{c|ccc}
         & MuTual \\
         \hline
         \# Context-Response Pairs & 8,860\\
         \# Avg. Turns per Dialogue & 4.73\\
         \# Avg. Words per Utterance & 19.57\\
         Vocabulary Size (Context) & 8,809 \\
         Vocabulary Size (Response) & 8,943 \\
         Vocabulary Size & 11,343 \\
         \# Original Dialogues & 6,371 \\
         \# Original Questions & 11,323 \\
         \# Response Candidates & 4\\
         \hline
    \end{tabular}
    \caption{Data statistics of MuTual.}
    \label{table:statistic}
     \vspace{-10pt}
\end{table}

The detailed statistics of MuTual are summarized in Table~\ref{table:statistic}.
MuTual has an average of 4.73 turns.
The vocabulary size is 11,343, which is smaller than other dialogue datasets \cite{ubuntu-dataset, douban-dataset}. Because MuTual is modified from listening tests of English as a foreign language, the complexity of morphology and grammar is much simpler than other datasets.

For human-annotated datasets, there is always a trade-off between the number of instances being annotated and the quality of annotations \cite{kryciski2019neural}. 
Our dataset is smaller than the previous crawling-based dialogue dataset \cite{ubuntu-dataset, douban-dataset} due to the collection method. But it is comparable with high-quality reasoning based dataset \cite{ai2-dataset, MRC-Multiple-sentence,commonsenseqa} and human-designed dialogue dataset \cite{personalizing-dataset}. Moreover, around 10k is sufficient to train a discriminative model \cite{Universal-Dependencies} or fine-tuning the pre-training model \cite{glue}.

To assess the distribution of different reasoning types, we annotate the specific types of reasoning that are involved for instance, sampled from the test set and categorize them into six groups. 
The definition and ratio of each group are shown as follows.

\textbf{Attitude Reasoning:}
This type of instance tests if a model knows the speaker's attitude towards an object. 

\textbf{Algebraic Reasoning:}
This type of instances tests whether a model is equipped with algebraic abilities when it chooses a response.

\textbf{Intention Prediction:}
This type tests whether a model can predict what the speaker is going to do next.

\textbf{Situational Reasoning:} Situation information (e.g., Location, Relationship between two speakers) is considered in this type of instance. A model should mine the implicit information from the previous context.

\textbf{Multi-fact Reasoning:}
In this type of instance, the correct response is related to multiple facts in context, which requires the model to deeply understand the context rather than simply text matching.

\textbf{Others:}. There are 9$\%$ of instances that require other commonsense knowledge. For example, at the bottom of Figure~\ref{figure:case}, the model should know that a fully reserved restaurant is usually very popular.

The six types of reasoning are considered the most relevant to real chatbots. For example, it enables chatbots to make personal recommendations if a machine knows the user's attitude. The ability of intention prediction allows chatbots to respond more intelligently in a long conversation session.

\subsection{MuTual$^{\text{plus}}$}
To further increase the difficulty, we use {\it safe response} to replace one of the candidate responses for each instance in \dataname.
To guarantee diversity, the safe response is sampled from a list including ``I'm afraid I didn't quite catch what you were saying.'', ``Could you repeat that?'', ``I'm really sorry, I didn't catch that.'', etc. In particular, once the instance is chosen, we randomly select a response to replace. If the positive response is replaced, the correct one is the safe response. If the negative response is replaced, the original positive response is still the best one.

The motivation to build MuTual$^{\text{plus}}$ is to evaluate whether a model is able to select a safe response when the other candidates are inappropriate. When we replace the positive response with a safe response, it simulates a scenario in which all the other candidates are incorrect. The phenomenon is common in retrieval-based chatbots, because limited candidate responses cannot handle all cases in practice.  
 Similarly, we can evaluate if the model can choose the correct response instead of a safe response when a correct response exists.

\section{Experiments}


We split the data into training, development and test sets, with an 80\%, 10\% and 10\% ratio. We pack instances constructed from the same conversation during splitting to avoid data leakage. 
Following the standard dialogue setting \cite{ubuntu-dataset, douban-dataset}, we consider our task as a response selection task and employ traditional information retrieval evaluation methods, including recall at position $1$ in 4 candidates (R$@1$), recall at position $2$ in 4 candidates (R$@2$) and Mean Reciprocal Rank (MRR) \cite{MRR}.  
We compare the performance of several response selection models as well as pre-training models. We simply introduce these works as follows:

\subsection{Baselines}
\label{sec:baseline}
We evaluate individual scoring methods, multi-choice methods and human performance in our experiment. Given a context $c$ and four candidates $(r_1,r_2,r_3,r_4)$, the individual scoring method computes a score for each choice independently with a score $g(c,r_i)$, and selects the individual with the highest score among four candidates. On the contrary, the multi-choice method selects the best one by classification over all choices, formulated as $h(c,r_1,r_2,r_3,r_4)$.

\textbf{TF-IDF}:
The correct response tends to share more words with the context than the incorrect ones. Following \citet{ubuntu-dataset}, we calculate the TF-IDF vectors for the context and each of the candidate responses, respectively, and then select the highest cosine similarity between the context and the candidate response as the model output. The ``IDF'' is calculated only on the training set.

\textbf{Dual LSTM}  \cite{ubuntu-dataset}:
Two LSTMs are used to encode context and response, respectively. The relevance between context and response is calculated by the similarity of the final hidden state from both LSTMs.

\begin{table*}[t] \small
    \centering
    \begin{tabular}{c|c|ccc|ccc} 
    \hline
 & & \multicolumn{3}{c|}{Dev} & \multicolumn{3}{c}{Test}  \\
    \hline
    Baseline category  & Baseline method & R$@1$ & R$@2$ & MRR & R$@1$ & R$@2$ & MRR \\
    \hline
  \multirow{2}{*}{Baseline}    &     Human & - & - & - & 0.938 & 0.971 & 0.964\\
    &    Random & 0.250 & 0.500 & 0.604 & 0.250 & 0.500 & 0.604\\
         \hline
     \multirow{6}{*}{\shortstack[c]{Individual scoring method\\ (discrimination)}}  &    
            TF-IDF & 0.276 & 0.541 & 0.541 & 0.279 & 0.536 & 0.542\\
      & Dual LSTM \cite{ubuntu-dataset} & 0.266 & 0.528 & 0.538 & 0.260 & 0.491 & 0.743\\
      &   SMN \cite{douban-dataset} & 0.274 & 0.524 & 0.575 & 0.299 & 0.585 & 0.595 \\
      &   DAM \cite{dam-model} & 0.239 & 0.463 & 0.575 & 0.241 & 0.465 & 0.518\\

      &   BERT \cite{bert} & 0.657 & 0.867 & 0.803 & 0.648 & 0.847 & 0.795\\ 
      &   RoBERTa \cite{roberta} & {\bf 0.695} & 0.878 & 0.824 & {\bf 0.713} & {\bf 0.892} & {\bf 0.836}\\
    \hline
     \multirow{2}{*}{\shortstack[c]{Individual scoring method\\ (generation)}}    &  GPT-2 \cite{gpt2} &  0.335 & 0.595 & 0.586 & 0.332 & 0.602 & 0.584\\
      &  GPT-2-FT \cite{gpt2} & 0.398 & 0.646 & 0.628 & 0.392 & 0.670 & 0.629\\            
      \hline
    \multirow{2}{*}{Multi-choice method}
      &   BERT-MC \cite{bert} & 0.661 & 0.871 & 0.806 & 0.667 & 0.878 & 0.810 \\
      &  RoBERTa-MC \cite{roberta} & 0.693 & {\bf 0.887} & {\bf 0.825} & 0.686 & 0.887 & 0.822\\
    \hline
    \end{tabular}
    \caption{Comparison of varying approaches on \dataname.}
    \label{tab:performance}
    \vspace{-5pt}
\end{table*}

\begin{table*}[t] \small
    \centering
    \begin{tabular}{c|c|ccc|ccc} 
    \hline
 & & \multicolumn{3}{c|}{Dev} & \multicolumn{3}{c}{Test}  \\
    \hline
    Baseline category  & Baseline method & R$@1$ & R$@2$ & MRR & R$@1$ & R$@2$ & MRR \\
    \hline
  \multirow{2}{*}{Baseline}    &     Human & - & - & - & 0.930 & 0.972 & 0.961\\
    &    Random & 0.250 & 0.500 & 0.604 & 0.250 & 0.500 & 0.604\\
         \hline
     \multirow{6}{*}{\shortstack[c]{Individual scoring method\\ (discrimination)}}  &    
            TF-IDF & 0.283 & 0.530 & 0.763 & 0.278 & 0.529 & 0.764\\
      &   SMN \cite{douban-dataset} & 0.264 & 0.524 & 0.578 & 0.265 & 0.516 & 0.627 \\
      &   DAM \cite{dam-model} & 0.261 & 0.520 & 0.645 & 0.272 & 0.523 & 0.695\\

      &   BERT \cite{bert} & 0.514 & 0.787 & 0.715 & 0.514 & 0.787 & 0.715\\ 
      &   RoBERTa \cite{roberta} & 0.622 & 0.853 & 0.782 & 0.626 & 0.866 & 0.787\\
    \hline
     \multirow{2}{*}{\shortstack[c]{Individual scoring method\\ (generation)}}    &  GPT-2 \cite{gpt2} &  0.305 & 0.565 & 0.562 & 0.316 & 0.574 & 0.568\\
      &  GPT-2-FT \cite{gpt2} & 0.226 & 0.577 & 0.528 & 0.226 & 0.611 & 0.535\\            
      \hline
    \multirow{2}{*}{Multi-choice method}
      &   BERT-MC \cite{bert} & 0.586 & 0.791 & 0.751 & 0.580 & 0.792 & 0.749 \\
      &  RoBERTa-MC \cite{roberta} & {\bf 0.621} & {\bf 0.830} & {\bf 0.778} & {\bf 0.643} & {\bf 0.845} & {\bf 0.792}\\
    \hline
    \multirow{2}{*}{Transfer method} & RoBERTa \cite{roberta} & 0.559 & 0.827 & 0.746 & 0.558 & 0.827 & 0.746\\
    & RoBERTa-MC \cite{roberta} & 0.384 & 0.815 & 0.656 & 0.402 & 0.845 & 0.673 \\
    \hline
    
    \end{tabular}
    \caption{Results on \dataname$^{\text{plus}}$. Transfer method denotes that we train it on \dataname\ and test on \dataname$^{\text{plus}}$.}
    \label{tab:plus}
    \vspace{-5pt}
\end{table*}

\textbf{Sequential Matching Network} \cite{douban-dataset}:
To avoid losing information in the context, SMN constructs a word-word and a sequence-sequence similarity matrix, instead of utilizing the last hidden state only, and then aggregates similarity matrix as a matching score.

\textbf{Deep Attention Matching Network}: \citet{dam-model} adopt self attention module \cite{transformer} to encode response and each utterance, respectively. To match utterance and response, DAM further applies cross-attention module and 3D matching to obtain final score.

\textbf{BERT} \cite{bert}: Pre-training models have shown promising results on various  multi-choice and reasoning tasks \cite{bert-result, bert-rc}. Following \citet{bert}, we concatenate the context (sentence A), and a candidate response (sentence B) as BERT input. 
On the top of BERT, a fully-connected layer is used for transforming the [CLS] token representation to the matching score.

\textbf{RoBERTa}:  \citet{roberta} re-establish BERT's masked language model training objective by using more data and different hyper-parameters. We fine-tune RoBERTa in the same way as BERT.

\textbf{GPT-2} \cite{gpt2}: Given a context, the positive response has a higher probability compared with negative responses. Motivated by this, we concatenate context and response as a sequence, and calculate the joint probability of an entire sequence. The response in the lowest perplexity sequence is considered as the positive response. Moreover, we fine-tune the GPT-2 on [Context, Positive Response] pairs in \dataname\ training set, denoted as \textbf{GPT-2-FT}. 

\textbf{Multi-choice Method}: Inspired by BERT for multiple choice \cite{bert}, the task is considered as picking the most suitable response by comparing four candidates responses.
In particular, we concatenate each candidate response with the corresponding context. 
Each input sequence is subsequently encoded to produce a [CLS] representation. The positive response is predicted based on the concatenation of all [CLS] representations, on which a fully connected layer with softmax is used.  
The method is denoted as \textbf{BERT-MC}.  Similarly, we implement \textbf{RoBERTa-MC} as another multi-choice method. 

\textbf{Human Performance}:
To obtain the human performance, we employ 3 NLP experts to measure the ceiling performance on the test set. 

\subsection{Experiment Results}

We report the performance of approaches introduced in \ref{sec:baseline}, and human performance. 
Implementation details are shown in Appendix B.
\subsubsection{Results on \dataname}
All models perform significantly worse than on other popular conversation datasets, such as the Ubuntu Corpus \cite{ubuntu-dataset} and the Dialogue NLI dataset \cite{dialogue-nli}, while human can address the reasoning problems easily. 
For example, BERT gives 85.8 \% R$_{10}@1$ on the Ubuntu Corpus, but RoBERTa only gives 71.3\% R$_4@1$ on \dataname.

TF-IDF only slightly better than randomly guessing, which indicates that there is no obvious statistic clue between context and positive response. In contrast, TF-IDF achieves 54.98$\%$ R$@1$ score on the Ubuntu Corpus, showing our dataset is more difficult to get the correct answer by text overlap. 
We evaluate typical retrieved-based dialogue models' performance on \dataname. From Table \ref{tab:performance}, we can see that well-designed matching models do not give better performance compared with simple dual LSTM, moreover, they drop by more than 50 absolute R$@1$ points compared to their performance on the Ubuntu Corpus, indicating that text matching models cannot handle reasoning problem well.

Both BERT and RoBERTa outperform other models in \dataname, which is consistent with results in other literatures \cite{commonsenseqa}.  This is mainly because models learn reasoning capability during the pre-training on a large corpus. Although RoBERTa only gets 71.3\% on R$@1$, it achieves a surprising number, 89.2 \%, on R$@2$, indicating that the model is able to rank the correct response to the top-2 position. BERT-MC and RoBERTa-MC obtain similar results with BERT and RoBERTa, respectively.
However, even RoBERTa is far behind human performance 23 points on R$@1$, indicating that \dataname\ is indeed a challenging dataset, which opens the door for tackling new and complex reasoning problems in multi-turn conversations.

GPT-2 and GPT-2-FT also perform undesirably on \dataname, even if the averaged perplexity on \dataname\ testset is 10.40. This phenomenon illustrates that 1) sentences in \dataname\ are fluent; and 2) current generative models still have plenty of room to improve their reasoning ability.

\subsubsection{Results on \dataname$^{\text{plus}}$}
As shown in Table~\ref{tab:plus}, all models perform worse on \dataname$^{\text{plus}}$, indicating the dataset is more difficult than \dataname, which is consistent with our assumption. We find that the performance of multi-choice method is significantly better than individual scoring method. One possible explanation is that multi-choice methods consider candidates together, so they can distinguish whether or not the safe response is the best one. In contrast, individual scoring methods are not robust, and safe responses are easy to confuse methods in the training stage. Moreover, RoBERTa-MC outperforms others by a large margin, showing its outstanding performance on reasoning problems. 

Furthermore, we conduct a transfer experiment, in which models are trained on \dataname\ but tested on \dataname$^{\text{plus}}$ without fine-tuning. The experiment investigates whether the model handles safe responses well if they have never seen them in training corpus.
As shown in Table~\ref{tab:plus}, RoBERTa-MC and RoBERTa drops 24.1\% and 6.8\%, respectively, in the transfer setting, demonstrating the benefits of seeing safe responses during the training process. Moreover, the individual scoring RoBERTa outperforms RoBERTa-MC, showing that the individual scoring method is more robust, when the safe response is not fed during training.
\subsection{Discussion}

\begin{figure}[t]
    \centering
    \includegraphics[width=0.4\textwidth]{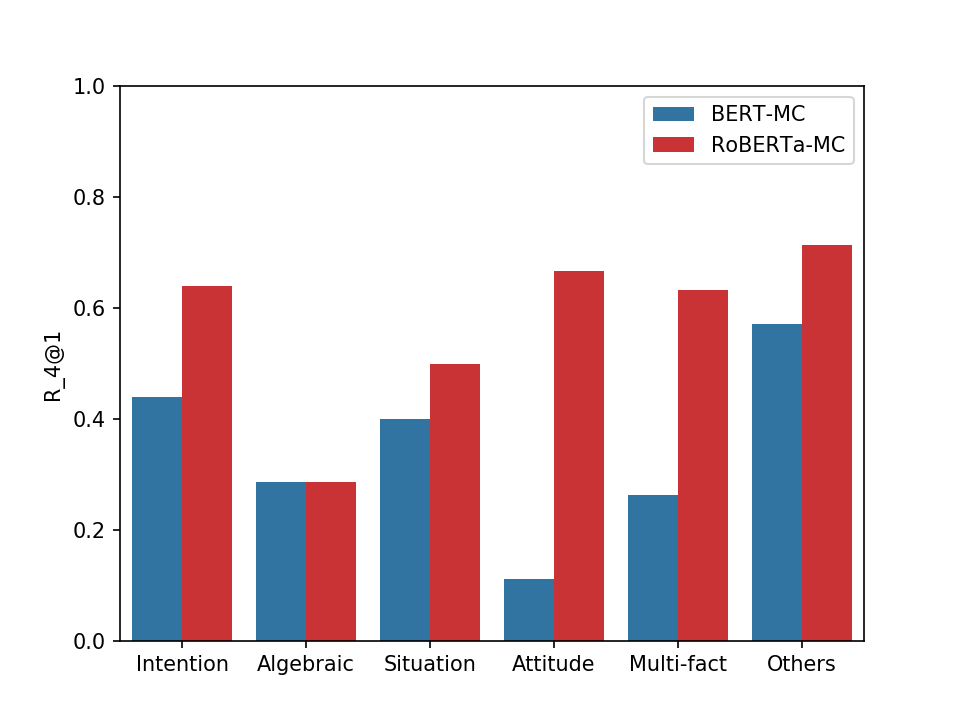}
    \caption{BERT-MC and RoBERTa-MC performance on different reasoning types.}
    \label{fig:reasoning-analysis}
\end{figure}

\begin{figure}
    \centering
    \includegraphics[width=0.5\textwidth]{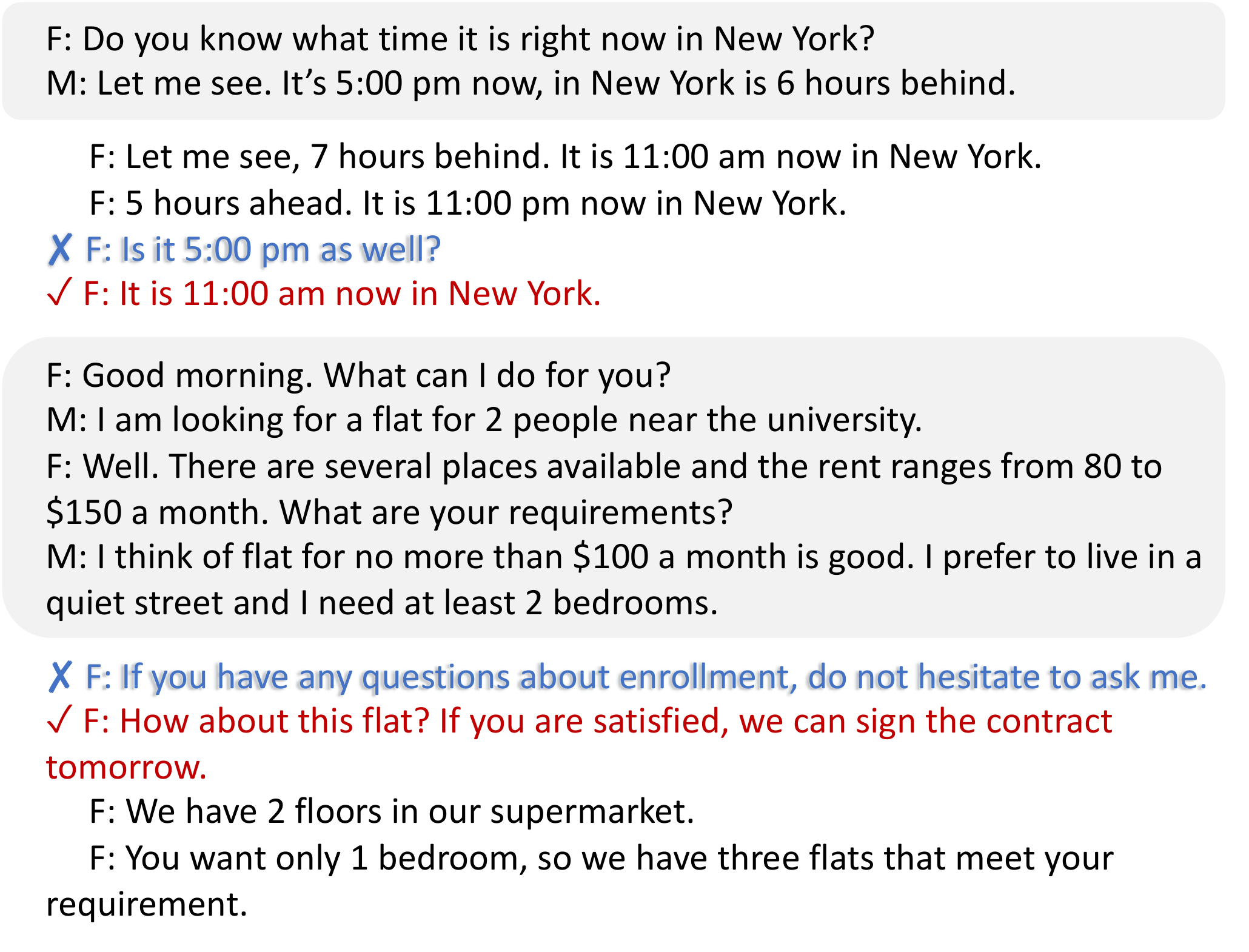}
    \caption{Error analysis. {\textcolor[rgb]{0.26,0.45,0.77} \XSolidBrush} indicates RoBERTa-MC's prediction.}
    \label{fig:case-study}
    \vspace{-10pt}
\end{figure}

{\bf Performance across different reasoning types:}
To analyze model performance across different reasoning types, we calculate BERT-MC and RoBERTa-MC performance on various question types as we introduce in Section~\ref{sec:data analysis}. 
As shown in Figure~\ref{fig:reasoning-analysis}, we find that the trends of BERT-MC and RoBERTa-MC are similar across different categories. RoBERTa-MC significantly outperforms BERT-MC in attitude reasoning and multi-fact reasoning. One potential reason is that there are some normal patterns between action and attitude captured by RoBERTa-MC, such as ``play football'' and ``excited''. However, instances that involve algebraic and situation show poor performance. These two reasoning types heavily depend on commonsense reasoning. Taking Figure~\ref{fig:case-study} as examples, it takes a simple subtraction step to derive the time difference (5:00 pm - 6h = 11:00 am), but this turns out a significant challenge for RoBERTa-MC. In the second case, RoBERTa-MC fails to infer the dialogue situation, where the goal is to find a flat to rent.

\begin{figure}[t!] \small
    \centering
    \includegraphics[width=0.4\textwidth]{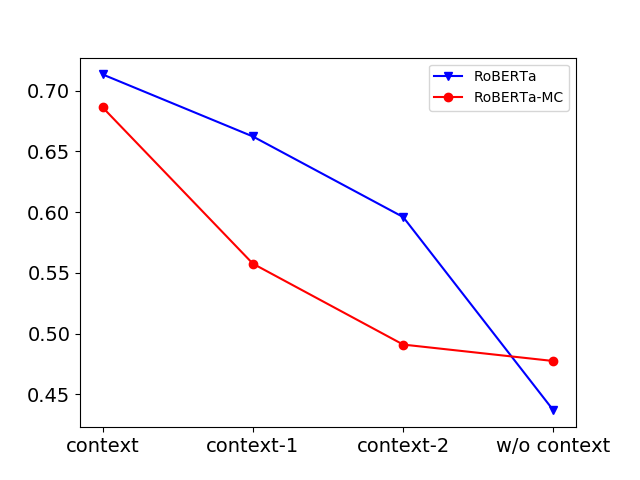}
    \caption{Ablation of context information. w/o context means all contexts are removed, so models just predict correct choice based on four candidates. context-n denotes the earlist n utterances are removed.}
    \label{fig:turns}
\end{figure}


{\bf Performance across different context lengths:}
It is interesting that the performance of RoBERTa does not decrease significantly with the number of turns increasing, which is different from the phenomenon observed on other datasets. As shown in Table~\ref{tab:context-length}, the performance drops by only 1.9 points R$@1$ from 2 turns to long turns ($>$6), and the performance of 5 turns is higher than those with 4 turns, indicating the reasoning problems do not become much harder when the context becomes longer. The results also show that the difficulty of \dataname\ is attributed to reasoning instead of complex conversation history.

\begin{table}[t!] \small
    \centering
    \begin{tabular}{c|ccccc}
    \hline
         & \#T=2 & \#T=3 & \#T=4 & \#T=5 &\#T$>$ 6 \\
    \hline
        \#Instances & 290 & 143 & 115 & 51 & 287 \\
         RoBERTa & 0.731 & 0.657 & 0.635 & 0.804 & 0.712 \\
         RoBERTa-MC & 0.681 & 0.622 & 0.609 & 0.725 & 0.750 \\
    \hline
    \end{tabular}
    \caption{Performance comparison (R$@1$) of different number of turns on the test set. \#T denotes number of turns.  \#Instances is the number of instances}
    \label{tab:context-length}
    \vspace{-10pt}
\end{table}

{\bf Context ablation study:} 
We further verify whether our dataset requires multi-turn understanding rather than degenerating to a single turn reasoning problem. We evaluate Roberta and Roberta-MC performance when some utterances are manually removed. Figure~\ref{fig:turns} shows the performance when the earliest $n$ utterances are removed in testing. As the ablation utterance increases, the performance of RoBERTa and RoBERTa-MC significantly decreases, which conforms to intuition. RoBERTa and RoBERTa-MC achieve only 43.7\% and 47.7\% after ablating all utterances in the context, respectively, indicating the importance of each utterance and the quality of the dataset. Moreover, if we shuffle the sequence of utterance, the performance of RoBERTa-MC drops by 3.8\% only, showing that it is insensitive to the utterance sequence information.

\section{Conclusion}
We introduced \dataname, a high-quality manually annotated multi-turn dialogue reasoning dataset, which contains 8,860 dialogues and aims to test reasoning ability of dialogue models. We describe the process for generating \dataname, and perform a detailed analysis. We find that various state-of-the-art models show poor performance in \dataname. The best model RoBERTa only obtains 71.3\% R$@1$. There is a large gap between the model performance and human performance. We hope that this dataset facilitates future research on multi-turn conversation reasoning problem.

\section*{Acknowledgments}

We thank Yulong Chen, Duyu Tang, Zhiyang Teng and Sen Yang for their insightful discussions. We also thank all anonymous reviewers for their constructive comments. The corresponding author is Yue Zhang. We thank the support by a BrightDreams Robotics - Westlake University research grant.

\bibliography{acl2020}

\begin{thebibliography}{37}
\expandafter\ifx\csname natexlab\endcsname\relax\def\natexlab#1{#1}\fi

\bibitem[{Nivre~et al.(2019)}]{Universal-Dependencies}
Joakim Nivre~et al. 2019.
\newblock \href {http://hdl.handle.net/11234/1-2988} {Universal dependencies
  2.4}.
\newblock {LINDAT}/{CLARIN} digital library at the Institute of Formal and
  Applied Linguistics ({{\'U}FAL}), Faculty of Mathematics and Physics, Charles
  University.

\bibitem[{Chen et~al.(2017)Chen, Zhu, Ling, Wei, Jiang, and Inkpen}]{esim}
Qian Chen, Xiaodan Zhu, Zhen-Hua Ling, Si~Wei, Hui Jiang, and Diana Inkpen.
  2017.
\newblock \href {https://doi.org/10.18653/v1/P17-1152} {Enhanced {LSTM} for
  natural language inference}.
\newblock In \emph{Proceedings of the 55th Annual Meeting of the Association
  for Computational Linguistics (Volume 1: Long Papers)}, pages 1657--1668,
  Vancouver, Canada. Association for Computational Linguistics.

\bibitem[{Clark et~al.(2019)Clark, Khandelwal, Levy, and
  Manning}]{bert-attention}
Kevin Clark, Urvashi Khandelwal, Omer Levy, and Christopher~D. Manning. 2019.
\newblock \href {https://doi.org/10.18653/v1/W19-4828} {What does {BERT} look
  at? an analysis of {BERT}{'}s attention}.
\newblock In \emph{Proceedings of the 2019 ACL Workshop BlackboxNLP: Analyzing
  and Interpreting Neural Networks for NLP}, pages 276--286, Florence, Italy.
  Association for Computational Linguistics.

\bibitem[{Clark et~al.(2018)Clark, Cowhey, Etzioni, Khot, Sabharwal, Schoenick,
  and Tafjord}]{ai2-dataset}
Peter Clark, Isaac Cowhey, Oren Etzioni, Tushar Khot, Ashish Sabharwal, Carissa
  Schoenick, and Oyvind Tafjord. 2018.
\newblock \href {http://arxiv.org/abs/1803.05457} {Think you have solved
  question answering? try arc, the ai2 reasoning challenge}.

\bibitem[{Devlin et~al.(2019)Devlin, Chang, Lee, and Toutanova}]{bert}
Jacob Devlin, Ming-Wei Chang, Kenton Lee, and Kristina Toutanova. 2019.
\newblock \href {https://doi.org/10.18653/v1/N19-1423} {{BERT}: Pre-training of
  deep bidirectional transformers for language understanding}.
\newblock In \emph{Proceedings of the 2019 Conference of the North {A}merican
  Chapter of the Association for Computational Linguistics: Human Language
  Technologies, Volume 1 (Long and Short Papers)}, pages 4171--4186,
  Minneapolis, Minnesota. Association for Computational Linguistics.

\bibitem[{Dua et~al.(2019)Dua, Wang, Dasigi, Stanovsky, Singh, and
  Gardner}]{drop}
Dheeru Dua, Yizhong Wang, Pradeep Dasigi, Gabriel Stanovsky, Sameer Singh, and
  Matt Gardner. 2019.
\newblock \href {https://doi.org/10.18653/v1/N19-1246} {{DROP}: A reading
  comprehension benchmark requiring discrete reasoning over paragraphs}.
\newblock In \emph{Proceedings of the 2019 Conference of the North {A}merican
  Chapter of the Association for Computational Linguistics: Human Language
  Technologies, Volume 1 (Long and Short Papers)}, pages 2368--2378,
  Minneapolis, Minnesota. Association for Computational Linguistics.

\bibitem[{Huang et~al.(2019)Huang, Le~Bras, Bhagavatula, and Choi}]{Cosmos}
Lifu Huang, Ronan Le~Bras, Chandra Bhagavatula, and Yejin Choi. 2019.
\newblock \href {https://doi.org/10.18653/v1/D19-1243} {Cosmos {QA}: Machine
  reading comprehension with contextual commonsense reasoning}.
\newblock In \emph{Proceedings of the 2019 Conference on Empirical Methods in
  Natural Language Processing and the 9th International Joint Conference on
  Natural Language Processing (EMNLP-IJCNLP)}, pages 2391--2401, Hong Kong,
  China. Association for Computational Linguistics.

\bibitem[{Jawahar et~al.(2019)Jawahar, Sagot, and Seddah}]{bert-analysis}
Ganesh Jawahar, Beno{\^\i}t Sagot, and Djam{\'e} Seddah. 2019.
\newblock \href {https://doi.org/10.18653/v1/P19-1356} {What does {BERT} learn
  about the structure of language?}
\newblock In \emph{Proceedings of the 57th Annual Meeting of the Association
  for Computational Linguistics}, pages 3651--3657, Florence, Italy.
  Association for Computational Linguistics.

\bibitem[{Kannan et~al.(2016)Kannan, Kurach, Ravi, Kaufmann, Tomkins, Miklos,
  Corrado, Luk{\'{a}}cs, Ganea, Young, and Ramavajjala}]{smartreply}
Anjuli Kannan, Karol Kurach, Sujith Ravi, Tobias Kaufmann, Andrew Tomkins,
  Balint Miklos, Greg Corrado, L{\'{a}}szl{\'{o}} Luk{\'{a}}cs, Marina Ganea,
  Peter Young, and Vivek Ramavajjala. 2016.
\newblock Smart reply: Automated response suggestion for email.
\newblock In \emph{Proceedings of the 22nd {ACM} {SIGKDD} International
  Conference on Knowledge Discovery and Data Mining, San Francisco, CA, USA,
  August 13-17, 2016}, pages 955--964.

\bibitem[{Khashabi et~al.(2018)Khashabi, Chaturvedi, Roth, Upadhyay, and
  Roth}]{MRC-Multiple-sentence}
Daniel Khashabi, Snigdha Chaturvedi, Michael Roth, Shyam Upadhyay, and Dan
  Roth. 2018.
\newblock \href {https://doi.org/10.18653/v1/N18-1023} {Looking beyond the
  surface: A challenge set for reading comprehension over multiple sentences}.
\newblock In \emph{Proceedings of the 2018 Conference of the North {A}merican
  Chapter of the Association for Computational Linguistics: Human Language
  Technologies, Volume 1 (Long Papers)}, pages 252--262, New Orleans,
  Louisiana. Association for Computational Linguistics.

\bibitem[{Kryściński et~al.(2019)Kryściński, Keskar, McCann, Xiong, and
  Socher}]{kryciski2019neural}
Wojciech Kryściński, Nitish~Shirish Keskar, Bryan McCann, Caiming Xiong, and
  Richard Socher. 2019.
\newblock \href {http://arxiv.org/abs/1908.08960} {Neural text summarization: A
  critical evaluation}.

\bibitem[{Lai et~al.(2017)Lai, Xie, Liu, Yang, and Hovy}]{lai-race}
Guokun Lai, Qizhe Xie, Hanxiao Liu, Yiming Yang, and Eduard Hovy. 2017.
\newblock \href {https://doi.org/10.18653/v1/D17-1082} {{RACE}: Large-scale
  {R}e{A}ding comprehension dataset from examinations}.
\newblock In \emph{Proceedings of the 2017 Conference on Empirical Methods in
  Natural Language Processing}, pages 785--794, Copenhagen, Denmark.
  Association for Computational Linguistics.

\bibitem[{Levesque et~al.(2012)Levesque, Davis, and Morgenstern}]{wsc-dataset}
{Hector J.} Levesque, Ernest Davis, and Leora Morgenstern. 2012.
\newblock The winograd schema challenge.
\newblock In \emph{13th International Conference on the Principles of Knowledge
  Representation and Reasoning, KR 2012}, pages 552--561.

\bibitem[{Liu et~al.(2019)Liu, Ott, Goyal, Du, Joshi, Chen, Levy, Lewis,
  Zettlemoyer, and Stoyanov}]{roberta}
Yinhan Liu, Myle Ott, Naman Goyal, Jingfei Du, Mandar Joshi, Danqi Chen, Omer
  Levy, Mike Lewis, Luke Zettlemoyer, and Veselin Stoyanov. 2019.
\newblock \href {http://arxiv.org/abs/1907.11692} {Roberta: A robustly
  optimized bert pretraining approach}.

\bibitem[{Lowe et~al.(2015)Lowe, Pow, Serban, and Pineau}]{ubuntu-dataset}
Ryan Lowe, Nissan Pow, Iulian Serban, and Joelle Pineau. 2015.
\newblock \href {https://doi.org/10.18653/v1/W15-4640} {The {U}buntu dialogue
  corpus: A large dataset for research in unstructured multi-turn dialogue
  systems}.
\newblock In \emph{Proceedings of the 16th Annual Meeting of the Special
  Interest Group on Discourse and Dialogue}, pages 285--294, Prague, Czech
  Republic. Association for Computational Linguistics.

\bibitem[{Pan et~al.(2019)Pan, Bai, Wang, Zhou, and
  Liu}]{pan-etal-2019-improving}
Zhufeng Pan, Kun Bai, Yan Wang, Lianqiang Zhou, and Xiaojiang Liu. 2019.
\newblock \href {https://doi.org/10.18653/v1/D19-1191} {Improving open-domain
  dialogue systems via multi-turn incomplete utterance restoration}.
\newblock In \emph{Proceedings of the 2019 Conference on Empirical Methods in
  Natural Language Processing and the 9th International Joint Conference on
  Natural Language Processing (EMNLP-IJCNLP)}, pages 1824--1833, Hong Kong,
  China. Association for Computational Linguistics.

\bibitem[{Radford et~al.(2019)Radford, Wu, Child, Luan, Amodei, and
  Sutskever}]{gpt2}
Alec Radford, Jeff Wu, Rewon Child, David Luan, Dario Amodei, and Ilya
  Sutskever. 2019.
\newblock Language models are unsupervised multitask learners.

\bibitem[{Reddy et~al.(2019)Reddy, Chen, and Manning}]{coqa}
Siva Reddy, Danqi Chen, and Christopher~D. Manning. 2019.
\newblock \href {https://doi.org/10.1162/tacl_a_00266} {Coqa: A conversational
  question answering challenge}.
\newblock \emph{Transactions of the Association for Computational Linguistics},
  7:249–266.

\bibitem[{Shang et~al.(2015)Shang, Lu, and Li}]{weibo-dataset}
Lifeng Shang, Zhengdong Lu, and Hang Li. 2015.
\newblock \href {https://doi.org/10.3115/v1/P15-1152} {Neural responding
  machine for short-text conversation}.
\newblock In \emph{Proceedings of the 53rd Annual Meeting of the Association
  for Computational Linguistics and the 7th International Joint Conference on
  Natural Language Processing (Volume 1: Long Papers)}, pages 1577--1586,
  Beijing, China. Association for Computational Linguistics.

\bibitem[{Shum et~al.(2018)Shum, He, and Li}]{xiaoice}
Heung-yeung Shum, Xiao-dong He, and Di~Li. 2018.
\newblock \href {https://doi.org/10.1631/FITEE.1700826} {From eliza to xiaoice:
  challenges and opportunities with social chatbots}.
\newblock \emph{Frontiers of Information Technology {\&} Electronic
  Engineering}, 19(1):10--26.

\bibitem[{Speer et~al.(2016)Speer, Chin, and Havasi}]{conceptnet}
Robert Speer, Joshua Chin, and Catherine Havasi. 2016.
\newblock \href {http://arxiv.org/abs/1612.03975} {Conceptnet 5.5: An open
  multilingual graph of general knowledge}.
\newblock In \emph{AAAI Conference on Artificial Intelligence}.

\bibitem[{Sun et~al.(2019)Sun, Yu, Chen, Yu, Choi, and Cardie}]{dream}
Kai Sun, Dian Yu, Jianshu Chen, Dong Yu, Yejin Choi, and Claire Cardie. 2019.
\newblock \href {https://doi.org/10.1162/tacl_a_00264} {{DREAM}: A challenge
  data set and models for dialogue-based reading comprehension}.
\newblock \emph{Transactions of the Association for Computational Linguistics},
  7:217--231.

\bibitem[{Talmor et~al.(2019)Talmor, Herzig, Lourie, and
  Berant}]{commonsenseqa}
Alon Talmor, Jonathan Herzig, Nicholas Lourie, and Jonathan Berant. 2019.
\newblock \href {https://doi.org/10.18653/v1/N19-1421} {{C}ommonsense{QA}: A
  question answering challenge targeting commonsense knowledge}.
\newblock In \emph{Proceedings of the 2019 Conference of the North {A}merican
  Chapter of the Association for Computational Linguistics: Human Language
  Technologies, Volume 1 (Long and Short Papers)}, pages 4149--4158,
  Minneapolis, Minnesota. Association for Computational Linguistics.

\bibitem[{Tao et~al.(2019)Tao, Wu, Xu, Hu, Zhao, and Yan}]{IoI}
Chongyang Tao, Wei Wu, Can Xu, Wenpeng Hu, Dongyan Zhao, and Rui Yan. 2019.
\newblock \href {https://doi.org/10.18653/v1/P19-1001} {One time of interaction
  may not be enough: Go deep with an interaction-over-interaction network for
  response selection in dialogues}.
\newblock In \emph{Proceedings of the 57th Annual Meeting of the Association
  for Computational Linguistics}, pages 1--11, Florence, Italy. Association for
  Computational Linguistics.

\bibitem[{Vaswani et~al.(2017)Vaswani, Shazeer, Parmar, Uszkoreit, Jones,
  Gomez, Kaiser, and Polosukhin}]{transformer}
Ashish Vaswani, Noam Shazeer, Niki Parmar, Jakob Uszkoreit, Llion Jones,
  Aidan~N Gomez, \L~ukasz Kaiser, and Illia Polosukhin. 2017.
\newblock \href
  {http://papers.nips.cc/paper/7181-attention-is-all-you-need.pdf} {Attention
  is all you need}.
\newblock In I.~Guyon, U.~V. Luxburg, S.~Bengio, H.~Wallach, R.~Fergus,
  S.~Vishwanathan, and R.~Garnett, editors, \emph{Advances in Neural
  Information Processing Systems 30}, pages 5998--6008. Curran Associates, Inc.

\bibitem[{Voorhees(2000)}]{MRR}
Ellen Voorhees. 2000.
\newblock The trec-8 question answering track report.

\bibitem[{Wang et~al.(2019)Wang, Singh, Michael, Hill, Levy, and Bowman}]{glue}
Alex Wang, Amanpreet Singh, Julian Michael, Felix Hill, Omer Levy, and
  Samuel~R. Bowman. 2019.
\newblock In \emph{7th International Conference on Learning Representations,
  {ICLR} 2019, New Orleans, LA, USA, May 6-9, 2019}. OpenReview.net.
\newblock \href {https://openreview.net/group?id=ICLR.cc/2019/Conference}
  {[link]}.

\bibitem[{Welleck et~al.(2019)Welleck, Weston, Szlam, and Cho}]{dialogue-nli}
Sean Welleck, Jason Weston, Arthur Szlam, and Kyunghyun Cho. 2019.
\newblock \href {https://doi.org/10.18653/v1/P19-1363} {Dialogue natural
  language inference}.
\newblock In \emph{Proceedings of the 57th Annual Meeting of the Association
  for Computational Linguistics}, pages 3731--3741, Florence, Italy.
  Association for Computational Linguistics.

\bibitem[{Whang et~al.(2019)Whang, Lee, Lee, Yang, Oh, and Lim}]{bert-result}
Taesun Whang, Dongyub Lee, Chanhee Lee, Kisu Yang, Dongsuk Oh, and HeuiSeok
  Lim. 2019.
\newblock \href {http://arxiv.org/abs/1908.04812} {Domain adaptive training
  bert for response selection}.

\bibitem[{Wu et~al.(2019)Wu, Wu, Xing, Xu, Li, and Zhou}]{Wu_2019}
Yu~Wu, Wei Wu, Chen Xing, Can Xu, Zhoujun Li, and Ming Zhou. 2019.
\newblock \href {https://doi.org/10.1162/coli_a_00345} {A sequential matching
  framework for multi-turn response selection in retrieval-based chatbots}.
\newblock \emph{Computational Linguistics}, 45(1):163–197.

\bibitem[{Wu et~al.(2017)Wu, Wu, Xing, Zhou, and Li}]{douban-dataset}
Yu~Wu, Wei Wu, Chen Xing, Ming Zhou, and Zhoujun Li. 2017.
\newblock \href {https://doi.org/10.18653/v1/P17-1046} {Sequential matching
  network: A new architecture for multi-turn response selection in
  retrieval-based chatbots}.
\newblock In \emph{Proceedings of the 55th Annual Meeting of the Association
  for Computational Linguistics (Volume 1: Long Papers)}, pages 496--505,
  Vancouver, Canada. Association for Computational Linguistics.

\bibitem[{Xu et~al.(2019)Xu, Liu, Shu, and Yu}]{bert-rc}
Hu~Xu, Bing Liu, Lei Shu, and Philip Yu. 2019.
\newblock \href {https://doi.org/10.18653/v1/N19-1242} {{BERT} post-training
  for review reading comprehension and aspect-based sentiment analysis}.
\newblock In \emph{Proceedings of the 2019 Conference of the North {A}merican
  Chapter of the Association for Computational Linguistics: Human Language
  Technologies, Volume 1 (Long and Short Papers)}, pages 2324--2335,
  Minneapolis, Minnesota. Association for Computational Linguistics.

\bibitem[{Young et~al.(2018)Young, Cambria, Chaturvedi, Zhou, Biswas, and
  Huang}]{dialogue-commonsense}
Tom Young, Erik Cambria, Iti Chaturvedi, Hao Zhou, Subham Biswas, and Minlie
  Huang. 2018.
\newblock Augmenting end-to-end dialogue systems with commonsense knowledge.
\newblock In \emph{Proceedings of the Thirty-Second {AAAI} Conference on
  Artificial Intelligence, (AAAI-18), the 30th innovative Applications of
  Artificial Intelligence (IAAI-18), and the 8th {AAAI} Symposium on
  Educational Advances in Artificial Intelligence (EAAI-18), New Orleans,
  Louisiana, USA, February 2-7, 2018}, pages 4970--4977.

\bibitem[{Zellers et~al.(2018)Zellers, Bisk, Schwartz, and Choi}]{swag-dataset}
Rowan Zellers, Yonatan Bisk, Roy Schwartz, and Yejin Choi. 2018.
\newblock \href {https://doi.org/10.18653/v1/D18-1009} {{SWAG}: A large-scale
  adversarial dataset for grounded commonsense inference}.
\newblock In \emph{Proceedings of the 2018 Conference on Empirical Methods in
  Natural Language Processing}, pages 93--104, Brussels, Belgium. Association
  for Computational Linguistics.

\bibitem[{Zhang et~al.(2018{\natexlab{a}})Zhang, Dinan, Urbanek, Szlam, Kiela,
  and Weston}]{personalizing-dataset}
Saizheng Zhang, Emily Dinan, Jack Urbanek, Arthur Szlam, Douwe Kiela, and Jason
  Weston. 2018{\natexlab{a}}.
\newblock \href {https://doi.org/10.18653/v1/P18-1205} {Personalizing dialogue
  agents: {I} have a dog, do you have pets too?}
\newblock In \emph{Proceedings of the 56th Annual Meeting of the Association
  for Computational Linguistics (Volume 1: Long Papers)}, pages 2204--2213,
  Melbourne, Australia. Association for Computational Linguistics.

\bibitem[{Zhang et~al.(2018{\natexlab{b}})Zhang, Li, Zhu, Zhao, and
  Liu}]{e-commerce-dataset}
Zhuosheng Zhang, Jiangtong Li, Pengfei Zhu, Hai Zhao, and Gongshen Liu.
  2018{\natexlab{b}}.
\newblock \href {https://www.aclweb.org/anthology/C18-1317} {Modeling
  multi-turn conversation with deep utterance aggregation}.
\newblock In \emph{Proceedings of the 27th International Conference on
  Computational Linguistics}, pages 3740--3752, Santa Fe, New Mexico, USA.
  Association for Computational Linguistics.

\bibitem[{Zhou et~al.(2018)Zhou, Li, Dong, Liu, Chen, Zhao, Yu, and
  Wu}]{dam-model}
Xiangyang Zhou, Lu~Li, Daxiang Dong, Yi~Liu, Ying Chen, Wayne~Xin Zhao, Dianhai
  Yu, and Hua Wu. 2018.
\newblock \href {https://doi.org/10.18653/v1/P18-1103} {Multi-turn response
  selection for chatbots with deep attention matching network}.
\newblock In \emph{Proceedings of the 56th Annual Meeting of the Association
  for Computational Linguistics (Volume 1: Long Papers)}, pages 1118--1127,
  Melbourne, Australia. Association for Computational Linguistics.

\end{thebibliography}
\bibliographystyle{acl_natbib}
\end{document}